\documentclass{llncs}
\usepackage{llncsdoc}
\usepackage{graphicx}
\usepackage{epsfig}
\usepackage{latexsym}

\newcommand{\argmin}{\mathop{min}\limits}
\newcommand{\argmax}{\mathop{argmax}\limits}

\begin{document}

\title{Distortion and Faults in Machine Learning Software\thanks{
Presented at the 9th SOFL+MSVL Workshop in Shenzhen, November 2019.}}


\author{Shin NAKAJIMA} 
\institute{National Institute of Informatics \\ Tokyo, Japan}

\maketitle

\begin{abstract}
Machine learning software, deep neural networks (DNN) software in particular,
discerns valuable information from a large dataset, a set of data.
Outcomes of such DNN programs are dependent on the quality of both
learning programs and datasets. Unfortunately, the quality of datasets is difficult
to be defined, because they are just samples.
The quality assurance of DNN software is difficult, because resultant
trained machine learning models are unknown prior to its development,
and the validation is conducted indirectly in terms of prediction performance.
%
%
This paper introduces a hypothesis that faults in the learning programs
manifest themselves as distortions in trained machine learning models.
Relative distortion degrees measured with appropriate observer functions
may indicate that there are some hidden faults.
The proposal is demonstrated with example cases of the MNIST dataset.
\end{abstract}


\section{Introduction}
Machine learning software, deep neural networks (DNN) software, 
is based on inductive methods to discern valuable information
from a given vast amount of data \cite{deep}. 
The quality of such software
is usually viewed from predication performance that obtained
approximation functions exhibit for incoming data.
Functional behavior of the resultant inference functions is dependent on
trained learning models, which learning programs calculate
with training datasets as their input. 

The quality of DNN software is dependent on both the learning programs
and training datasets; either or both is a source of degraded quality.
The learning programs result in inappropriate trained learning models
if they are not implemented faithfully in regard to well-designed
machine learning algorithms \cite{haykin},
Moreover, problematic datasets, suffering from sample 
selection bias \cite{dataset2009} for example,
have negative impacts on trained learning models.

Although the learning programs and datasets are sources to affect the quality
of inference functions, they are more or less indirect.
Trained learning models determine the quality directly, but have not been 
considered as \emph{first-class~citizens} to study quality issues.
The models are important numeral data, but are intermediate in that
they are synthesized by learning programs and then
transferred to inference programs.

This paper adapts a hypothesis that 
distortions in the trained learning models manifest themselves as 
faults resulting in quality degradation. 
Although such distortion is difficult to be measured directly as they are,
relative distortion degrees can be defined.
Moreover, this paper proposes a new way of generating datasets that show
characteristics of the dataset diversity \cite{nkjm2018S}, 
which is supposed to be effective in testing machine learning software 
from various ways.

\section{Machine Learning Software} \label{sec-ml}

\subsection{Learning Programs}

We consider supervised machine learning classifiers using
networks of perceptrons \cite{haykin} or deep neural networks \cite{deep}.
Our goal is to synthesize, inductively from a large dataset,
an approximation input-output relation classifying a multi-dimensional vector
data $\vec{a}$ into one of $C$ categories. 
The given dataset $LS$ is a set of $N$ number of pairs,
${\langle}{\vec{x}}^{n},{t}^{n}{\rangle}$ $(n = 1,{\cdots},N)$,
where a supervisor tag ${t}^{n}$ takes a value of $c$ $(c{\in}[1,{\cdots},C])$.
A pair ${\langle}{\vec{x}}^{n},{t}^{n}{\rangle}$ in $LS$ means that
${\vec{x}}^{n}$ is classified as ${t}^{n}$.

Given a learning model ${{y}}({{W}};{\vec{x}})$ as a multi-dimensional
non-linear function, differentiable with respect to both learning parameters ${W}$
and input data ${\vec{x}}$. Learning aims at obtaining a set of learning parameter
values (${W}^{*}$) by solving a numerical optimization problem. 
%
\[
{W}^{*}~=~arg{\argmin_{W}}~{\cal E}(W;X), ~~~~{\cal E}(W;X)~=~\frac{1}{N}{\sum_{n=1}^{N}}~{\ell}({{y}}({W};~{\vec{x}}^{n}),~{\vec{t}}^{n})
\]

\noindent The function $\ell(\_,\_)$ denotes distances between its two parameters,
representing how much a calculated output
${{y}}({W};{\vec{x}}^{n})$ differs from its accompanying
supervisor tag value ${\vec{t}}^{n}$.

We denote a function to calculate ${W}^{*}$ as ${\cal L}_{f}(LS)$,
which is a program
to solve the above optimization problem with its input dataset $LS$.
Moreover, we denote the empirical distribution of $LS$ as ${\rho}_{em}$.
${W}^{*}$ is a collection of learning parameter values to minimize the
mean of $\ell$ under ${\rho}_{em}$.

From viewpoints of the software quality,
${\cal L}_{f}(LS)$, a learning program, is concerned with the product quality,
because it must be a faithful implementation of
a machine learning method, the supervised learning method for this case.

\subsection{Inference Programs}
We introduce another function ${\cal I}_{f}(\vec{x})$,
using a trained learning model ${W}^{*}$ or ${{y}}({{W}^{*}};{\vec{x}})$,
calculates inference results of an incoming data $\vec{x}$.

For classification problems, the inference results are often 
expressed as probability that the data $\vec{x}$ belongs to a category $c$.
${Prob}({W},\vec{a};c)$, a function of $c$,
 is probability such that the data $\vec{a}$ is
classified as $c$.
This ${Prob}$ is readily implemented in
${\cal I}_{f}(x)$ if we choose 
\emph{Softmax} as an activation function of the output layer
of the learning model.

The prediction performance of ${\cal I}_{f}(\vec{x})$ is, indeed,
defined compactly as the accuracy of classification results
for a dataset $TS$ different from $LS$ used to calculate ${W}^{*}$;
$TS = \{{\langle}{\vec{x}}^{m}, {\vec{t}}^{m}{\rangle}\}$.
For a specified ${W}$,
${Correct}$ is a set-valued function to obtain a subset of data vectors
in $TS$.

\[
{Correct}({W};TS)
{\equiv}~\{ ~{\vec{x}}^{m} ~{\mid}
~{c}^{*}={\argmax_{c{\in}[1,{\ldots},C]}}~{Prob}({W},{\vec{x}}^{m};{c})
~{\wedge}~{\vec{t}}^{m}={c}^{*} ~\}
\]

\noindent If we express ${\mid}~{S}~{\mid}$ as a size of a set $S$, then
an accuracy is defined as a ratio as below.

\[
{Accuracy}({W}; TS) = \frac{{\mid}~{Correct}({W};TS)~{\mid}}{{\mid}~{TS}~{\mid}}
\]

\noindent Given a ${W}^{*}$ obtained by ${\cal L}_{f}(LS)$, 
the predication performance of ${\cal I}_{f}(LS)$ is defined in terms of
${Accuracy}({W}^{*};TS)$ for a dataset $TS$ different from $LS$.
For an individual incoming data $\vec{a}$, a function of $c$ $Prob({W}^{*},\vec{a};c)$
is a good performance measure.


\subsection{Quality Issues} \label{subsec-acc}

\begin{figure}[!b]
\begin{center}
\includegraphics[width=4.0in]{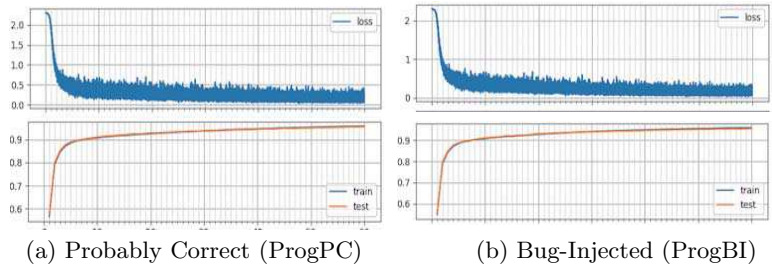}

{(a) Probably Correct (ProgPC)}\hspace*{1.5cm}{(b) Bug-Injected (ProgBI)}
\end{center}
\caption{Loss and Accuracy: MNIST Dataset} \label{fig-SGD}
\end{figure}

\subsubsection{Loss and Accuracy}

An NN learning problem is non-convex optimization, and thus
reaching a globally optimal solution is not guaranteed (e.g. \cite{haykin}).
The learning program ${\cal L}_{f}(LS)$ is iterated over epochs to search for
solutions and is taken as converged when the value of the loss (${\cal E}(W;LS)$)
is not changed much between two consecutive epochs.
The learning parameter values at this converged epoch are taken as ${W}^{*}$.
The derived ${W}^{*}$ may not be optimal, and thus an accuracy
is calculated to ensure that the obtained ${W}^{*}$ is appropriate.
Moreover,
${W}^{*}$ may be over-fitting to the training dataset $LS$.

In the course of the iteration, at an epoch $e$, the learning parameter
values ${W}^{(e)}$ are extracted.
The iteration is continued until the accuracy becomes satisfactory.
Both $Accuracy({W}^{(e)},LS)$ and $Accuracy({W}^{(e)},TS)$
are monitored to ensure the training process goes as desired.
If the accuracy of $TS$ is not much different from the accuracy with $LS$,
we may think that the learning result does not have the over-fitting problem.

Figure \ref{fig-SGD} shows loss and accuracy graphs as epochs proceed
measured during experiments\footnote{We return to the experiment later
in Section \ref{sec-case}.}.
The graphs, for example in Figure \ref{fig-SGD} (a),
 actually demonstrate that the search converges at
a certain desirable point in the solution space because the
loss decreases to be almost stable below a certain threshold,
and the accuracies of both $TS$ and $LS$ reach
a satisfactory level of higher than 0.95.
Figure \ref{fig-SGD} shows that the loss together with the accuracy
may be good indicators to decide whether NN training processes
behave well or not.

\subsubsection{Sources of Faults}
Intuitively, NN machine learning software shows good quality if 
prediction performance of ${\cal I}_{f}$ is acceptable. 
The graphs in Figure \ref{fig-SGD}, however, depict that
there is a counterexample case as discussed in \cite{nkjm2018S},
in which the learning task uses
MNIST dataset, for classifying hand-written numbers.

Figure \ref{fig-SGD}(a) are graphs of loss and accuracy of
a probably correct implementation of NN learning program,
while Figure \ref{fig-SGD}(b) are those of a bug-injected program.
The two graphs for loss are mostly the same to be converged.
The two accuracy graphs are similar as well, although
the program of Figure \ref{fig-SGD}(b) has faults in it.

MNIST dataset is a standard benchmark and is
supposed to be well-prepared free from any sample selection bias.
A bug-injected program ${\cal L}_{f}$ accepts
a training dataset $LS$ of MNIST and calculates a set of trained
learning parameters ${W}^{*}$. Intuitively, this ${W}^{*}$
is inappropriate, because a bug-injected program produces it.
However, the accuracy graphs show that there is no clear sign
of faults in the prediction results of ${\cal I}_{f}$, although its
behavior is completely determined by the probably inappropriate ${W}^{*}$.

A question arises how faults in ${\cal L}_{f}$ affect ${W}^{*}$,
which follows another question whether such \emph{faults} in ${W}^{*}$
are observable.

\section{Distortion Degrees} \label{sec-faults}

\subsection{Observations and Hypotheses} \label{subsec-preceq}


Firstly, we introduce a few new concepts and notations.
For two datasets ${DS}_{1}$ and ${DS}_{2}$,
a relation ${DS}_{1}~{\preceq}~{DS}_{2}$ denotes that
${DS}_{2}$ is more distorted than ${DS}_{1}$.
%
%
For two sets of trained learning parameters ${W}^{*}_{1}$ and ${W}^{*}_{2}$
of the same machine learning model ${y}({W}; \_)$,
a relation ${W}^{*}_{1}~{\preceq}~{W}^{*}_{2}$ denotes
that ${W}^{*}_{2}$ is more distorted than ${W}^{*}_{1}$.
A question here is how to measure such degrees of the distortion.
We assume a certain observer function $obs$ and a relation ${\cal R}_{obj}$
with a certain small threshold ${\epsilon}_{obs}$ such that
${\cal R}_{obj}~=~({\mid}obs({W}^{*}_{1}) - obs({W}^{*}_{2}){\mid}~{\le}~{\epsilon}_{obs})$.
The distortion relation is defined in terms of ${\cal R}_{obj}$,
${W}^{*}_{1}~{\preceq}~{W}^{*}_{2}~{\Leftrightarrow}~{\neg}{\cal R}_{obj}({W}^{*}_{1},{W}^{*}_{2})$.
We introduce below
three hypotheses referring to these notions.

\begin{quotation}
\noindent {\bf [{Hyp-1}]}~
Given a training dataset $LS$, a machine learning program
${\cal L}_{f}(LS)$, either correct or faulty, derives its optimal solution
${W}^{*}$. 
For the training dataset $LS$ and a testing dataset $TS$,
if both are appropriately sampled or both follows the same empirical distribution,
then $Accuracy({W}^{*}, TS)$ is almost the same as $Accuracy({W}^{*}, LS)$.
\end{quotation}

\begin{quotation}
\noindent {\bf [{Hyp-2}]}~
For a training program ${\cal L}_{f}$ and
two training datasets ${LS}_{j}$ ($j = 1$ or $2$),
if ${W}^{*}_{j}~=~{\cal L}_{f}({LS}_{j})$,
then ${LS}_{1}{\preceq}{LS}_{2}~{\Rightarrow}~{W}^{*}_{1}{\preceq}{W}^{*}_{2}$.
\end{quotation}

\begin{quotation}
\noindent {\bf [{Hyp-3}]}~
For two training datasets ${LS}_{j}$ ($j = 1$ or $2$) such that
${LS}_{1}{\preceq}{LS}_{2}$ and a certain appropriate $obs$,
if ${\cal L}_{f}$ is correct with respect to its functional specifications,
then two results, ${W}^{*}_{j}~=~{\cal L}_{f}({LS}_{j})$,
are almost the same, written as ${W}^{*}_{1}{\approx}{W}^{*}_{2}$
(or $obs({W}^{*}_{1}){\approx}obs({W}^{*}_{2})$).
However, if ${\cal L}_{f}^{\prime}$ is a faulty implementation,
then ${W}^{{\prime}*}_{1}{\prec}{W}^{{\prime}*}_{2}$.
\end{quotation}

\noindent The accuracy graph in Figure \ref{fig-SGD} is an instance of [{Hyp-1}].
In Figure \ref{fig-SGD} (a), the accuracy graphs for $TS$ and $LS$ are mostly
overlapped, and the same is true for the case of Figure \ref{fig-SGD} (b),
which illustrates that the accuracy is satisfactory even if
the learning programs is buggy.

Moreover,
the example in \cite{nkjm2018S} is an instance of the [{Hyp-2}] because of the followings.
A training dataset ${LS}^{(K+1)}$ is obtained 
by adding a kind of disturbance signal to ${LS}^{(K)}$ so that
${LS}^{(K)}{\preceq}{LS}^{(K+1)}$. With an appropriate observer
function ${obs}_{d}$, ${\cal R}_{{obs}_{d}}({W}^{(K)*},{W}^{(K+1)*})$
is falsified where ${W}^{(K)*} = {\cal L}_{f}({L}^{(K)})$.

\subsection{Generating Distorted Dataset} \label{subsec-generation}

We propose a new test data generation method.
We first explain the L-BFGS \cite{szegedy2014},
which illustrates a simple way to calculate adversarial examples.

Given
a dataset $LS$ of $\{{\langle}{\vec{x}}^{n},{\vec{t}}^{n}{\rangle}\}$,
${W}^{*}={\cal L}_{f}(LS)$. 
An adversarial example is a solution of
an optimization problem; 

\[
{\vec{x}}^{A} = arg{\argmin_{\vec{x}}}~{A}_{\lambda}({W}^{*};{\vec{x}}_{S},{{t}}_{T},\vec{x}),
\]
\[
{A}_{\lambda}({W}^{*};{\vec{x}}_{S},{{t}}_{T},{\vec{x}}) = {\ell}({{y}}({W}^{*};\vec{x}),{{t}}_{T})~+~{\lambda}{\cdot}{\ell}(\vec{x},{\vec{x}}_{S}).
\]

\noindent Such a data
${\vec{x}}^{A}$ is visually
close to a seed ${\vec{x}}_{S}$ for human eyes, 
but is actually added a faint noise so as to induce miss-inference
such that ${y}({W}^{*};{\vec{x}}^{A}) = {{t}}_{T}$.

Consider an optimization problem, in which a seed
${\vec{x}}_{S}$ is ${\vec{x}}^{n}$ and its target label ${{t}}_{T}$ is
${{t}}^{n}$.
\[
{\vec{x}}^{n*} = arg{\argmin_{\vec{x}}} ~A_{\lambda}({W}^{*};{\vec{x}}^{n},{{t}}^{n},\vec{x}).
\]
\noindent The method is equivalent to constructing
a new data ${\vec{x}}^{n*}$ to be added small noises.
Because the inferred label is not changed, ${\vec{x}}^{n*}$ is
not adversarial, but is \emph{distorted} from the seed ${\vec{x}}^{n}$.
When the value of the hyper-parameter $\lambda$ is chosen to be very small,
the distortion of ${\vec{x}}^{n*}$ is large from ${\vec{x}}^{n}$.
On the other hand, if $\lambda$ is appropriate, the effects of the noises
on ${\vec{x}}^{n*}$ can be small so that the data is close to the
original ${\vec{x}}^{n}$.

By applying the method to all the elements in $LS$, a new dataset is
obtained to be $\{ {\langle}{\vec{x}}^{n*},~{\vec{t}}^{n}{\rangle} \}$.
We introduce a function ${T}_{A}$ to generate such a dataset from $LS$ and ${W}^{*}$.
\[
{T}_{A}(\{{\langle}{\vec{x}}^{n},{\vec{t}}^{n}{\rangle}\},{W}^{*}) = \{~{\langle}arg{\argmin_{\vec{x}}} ~A_{\lambda}({W}^{*};{\vec{x}}^{n},{\vec{t}}^{n},\vec{x}),~{\vec{t}}^{n}{\rangle} ~\}.
\]
\noindent Now, ${LS}^{(K+1)} = {T}_{A}({LS}^{(K)},{\cal L}_{f}({LS}^{(K)}))$
(for $K{\ge}0$ and ${LS}^{(0)} = LS$).

\subsection{Some Properties} \label{subsec-property}

This section presents some properties that generated datasets ${LS}^{(K)}$
satisfy; ${LS}^{(K)} = {T}_{A}({LS}^{(K-1)},{\cal L}_{f}({LS}^{(K-1)}))$
where ${LS}^{(0)}$ is equal to be a given training dataset $LS$.

\begin{quotation}
\noindent {\bf [{Prop-1}]}~${LS}^{(K)}$ serves the same machine learning task
as $LS$ does.
\end{quotation}
\noindent We have that ${W}^{(0)*} = {\cal L}_{f}(LS)$.
As the optimization problem with ${W}^{(0)*}$ indicates, 
${\vec{x}}^{(n)*}$, an element of
${LS}^{(1)}$ does not deviate much from ${\vec{x}}^{(n)}$ in ${LS}$,
and is almost the same as ${\vec{x}}^{(n)}$ in special cases.
Therefore, ${LS}^{(1)}$ serves as the same machine learning task as $LS$ does.
Similarly, ${LS}^{(K)}$ serves as the same machine learning task as ${LS}^{(K-1)}$ does.
By induction, ${LS}^{(K)}$ serves as the same machine learning task 
as ${LS}$ does, although the deviation may be large. ${\Box}$

\begin{quotation}
\noindent {\bf [{Prop-2}]}~${LS}^{(K)}~{\preceq}~{LS}^{(K+1)}$
\end{quotation}
\noindent The distortion relation is satisfied by construction
if we take ${LS}^{(K)}$ as a starting criterion.${\Box}$

\begin{quotation}
\noindent {\bf [{Prop-3}]}~${LS}^{(K)}$ is more over-fitted to ${W}^{(K-1)*}$
than ${LS}^{(K-1)}$.
\end{quotation}
\noindent In the optimization problem, 
if the loss ${\ell}({\vec{y}}({W}^{(K-1)*}; {\vec{x}}^{*}),{t}_{T})$
is small, ${\vec{x}}^{*}$ in ${LS}^{(K)}$ can be considered to
be well-fitted to ${W}^{(K-1)*}$ because the data reconstruct
the supervisor tag ${t}_{T}$ well.
We make the loss is so small as in the above by choosing carefully
an appropriate $\lambda$ value.${\Box}$

\begin{quotation}
\noindent {\bf [{Prop-4}]}~There exists a certain ${K}_{c}$ such that,
for all $K$ to satisfy a relation ${K}{\ge}{K}_{c}$,
$Accuracy({W}^{(K)*},TS)~{\approx}~Accuracy({W}^{({K}_{c})*},TS)$
and
$Accuracy({W}^{(0)*},TS)~{\ge}~Accuracy({W}^{({K}_{c})*},TS)$.
$TS$ is a dataset different from $LS$, but follows
the same empirical distribution ${\rho}_{em}$.
\end{quotation}
\noindent From Prop-3, we can see ${LS}^{(K)}$ is over-fitted to
${W}^{({K}_{c})*}$ if $K = {K}_{c} + 1$. 
Because ${W}^{(K)*} = {\cal L}_{f}({LS}^{(K)})$
and both $LS$ and $TS$ follow the empirical distribution 
${\rho}_{em}$,
we have $Accuracy({W}^{(K)*},TS)~{\approx}~Accuracy({W}^{({K}_{c})*},TS)$.
Furthermore, ${LS}~{\preceq}~{LS}^{({K}_{c})}$ implies
$Accuracy({W}^{(0)*},LS)~{\ge}~Accuracy({W}^{({K}_{c})*},LS)$
and thus
$Accuracy({W}^{(0)*},TS)~{\ge}~Accuracy({W}^{({K}_{c})*},TS)$.${\Box}$

\begin{quotation}
\noindent {\bf [{Prop-5}]}~The dataset and trained learning model
reach respectively ${LS}^{\infty}$ and ${W}^{(\infty)*}$
if we repeatedly conduct the training ${\cal L}_{f}$
and the dataset generation ${T}_{A}$ interleavingly.
\end{quotation}
\noindent 
If we choose a $K$ to be sufficiently larger than ${K}_{c}$, 
we have, from Prop-4,
$Accuracy({W}^{(K)*},{LS}^{(K)})~{\approx}~Accuracy({W}^{({K}_{c})*},{LS}^{(K)})$,
which may imply that 
$Accuracy({W}^{(K)*},{LS}^{(K)})~{\approx}~Accuracy({W}^{(K+1)*},{LS}^{(K+1)})$.
From Prop-3, ${LS}^{(K+1)}$ is over-fitted to ${W}^{(K)*}$,
and thus we have ${LS}^{(K)}~{\approx}~{LS}^{(K+1)}$, which 
implies that we can choose a representative ${LS}^{(\infty)}$ from them.
Using this dataset, we have 
that ${W}^{(\infty)*} = {\cal L}_{f}({LS}^{(\infty)})$,
and that ${W}^{(\infty)*}$ is a representative. ${\Box}$


\section{A Case Study} \label{sec-case}

\subsection{MNIST Classification Problem}

MNIST dataset is a standard problem of classifying 
handwritten numbers \cite{mnist1998}.
It consists of a training dataset $LS$ of 60,000 vectors, and a testing dataset $TS$
of 10,000. Both $LS$ and $TS$ are randomly selected from a pool of vectors, and thus
are considered to follow the same empirical distribution.
The machine learning task is to classify an input sheet, or a vector data,
into one of ten categories from 0 to 9. A sheet is presented as
28$\times$28 pixels, each taking a value between 0 and 255 to represent
gray scales. Pixel values represent handwritten strokes, and a number
appears as a specific pattern of these pixel values.

In the experiments, the learning model is a classical neural network
with a hidden layer and an output layer. Activation function for neurons
in the hidden layer is \emph{ReLU}; its output is linear for positive input values
and a constant zero for negatives. 
A \emph{softmax} activation function is introduced so that
the inference program ${\cal I}_{f}$ returns probability that
an incoming data belongs to the ten categories.

\subsection{Experiments}

We prepared two learning programs ${\cal L}_{f}^{PC}$ and ${\cal L}_{f}^{BI}$.
The former is a probably correct implementation of a learning algorithm,
and the latter is a bug-injected version of the former.
We conducted two experiments in parallel, one using ${\cal L}_{f}^{PC}$
and the other with ${\cal L}_{f}^{BI}$, and made comparisons.
Below, we use notations such as ${\cal L}_{f}^{MD}$ where $MD$ is
either $PC$ or $BI$.

\subsubsection{Training with MNIST dataset}
We conducted trainings the MNIST training dataset ${LS}^{(0)}$;
${W}^{(0)*}_{MD}~=~{\cal L}_{f}^{MD}({LS}^{(0)})$.
Figure \ref{fig-SGD} illustrates several graphs to show their behavior, 
that are obtained in the training processes.
Both accuracy graphs in Figure \ref{fig-SGD} show that
$Accuracy({W}^{(0)*}_{MD},LS)$ and $Accuracy({W}^{(0)*}_{MD},TS)$ are
mostly the same. In addition, $Accuracy({W}^{(0)*}_{PC},\_)$ and
$Accuracy({W}^{(0)*}_{BI},\_)$ are indistinguishable.
The above observation is consistent with {\bf [Hyp-1]}.

\subsubsection{Generating Distorted Datasets}
We generated distorted datasets with the method described in 
Section \ref{subsec-generation}.
We introduce here short-hand notations such as ${LS}^{(K)}_{MD}$;
${LS}^{(1)}_{MD} = {T}_{A}({LS}^{(0)},{\cal L}_{f}^{MD}({LS}^{(0)}))$.

\begin{figure}[!t]
\begin{center}

\includegraphics[width=4.0in]{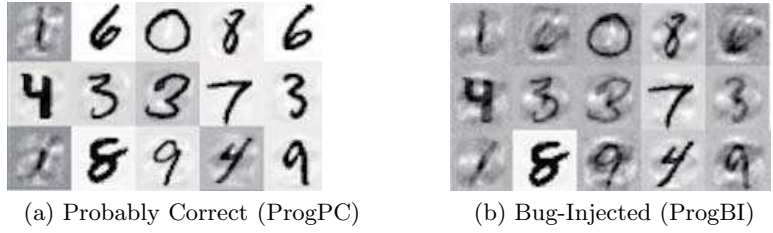}

{(a) Probably Correct (ProgPC)}\hspace*{1.5cm}{(b) Bug-Injected (ProgBI)}
\end{center}
\caption{Distorted Data} \label{fig-distorted}
\end{figure}

Figure \ref{fig-distorted} shows a fragment of ${LS}^{(1)}_{MD}$.
We recognize that all the data are not so clear as those of the original MNIST dataset
and thus are considered distorted. 
We may consider them as ${LS}~{\preceq}~{LS}^{(1)}_{MD}$, which is
an instance of {\bf [Prop-2]}.
Furthermore, for human eyes, 
Figure \ref{fig-distorted} (b) for the case with ${\cal L}_{f}^{BI}$
is more distorted than Figure \ref{fig-distorted} (a) of ${\cal L}_{f}^{PC}$,
which may be described as ${LS}^{(1)}_{PC}~{\preceq}~{LS}^{(1)}_{BI}$.


\subsubsection{Training with Distorted Datasets}
\begin{figure}[!t]
\begin{center}
\includegraphics[width=4.0in]{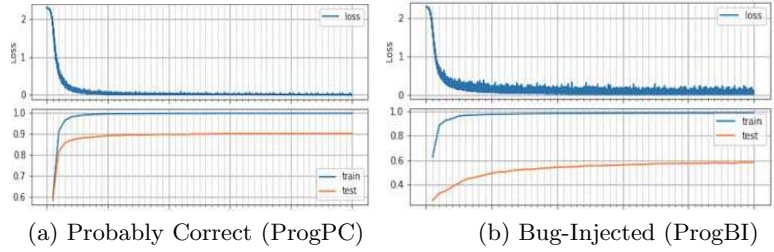}

{(a) Probably Correct (ProgPC)}\hspace*{1.5cm}{(b) Bug-Injected (ProgBI)}

\end{center}
\caption{Loss and Accuracy: Distorted Training Dataset} \label{fig-rel}
\end{figure}

We then conducted trainings the distorted dataset ${LS}^{(1)}_{MD}$;
${W}^{(1)*}_{MD}~=~{\cal L}_{f}^{MD}({LS}^{(1)}_{MD})$.
Figure \ref{fig-rel} shows loss and accuracy graphs in their learning processes.
Comparing Figure \ref{fig-rel} with Figure \ref{fig-SGD} leads to 
the following observations.

Firstly, the overall loss values of Figure \ref{fig-rel} are smaller 
than those of Figure \ref{fig-SGD} counterparts,
and the metrics concerning with the differences ($\ell(\_,\_)$)
are small for the distorted dataset cases.
%

Secondly, for the MNIST testing dataset $TS$,
$Accuracy({W}^{(1)*}_{MD},{TS})$ is lower than
$Accuracy({W}^{(0)*}_{MD},{TS})$, while
$Accuracy({W}^{(1)*}_{MD},{LS}^{(1)}_{MD})$ reaches close to $100{\%}$.
Together with the fact of ${LS}~{\preceq}~{LS}^{(1)}_{MD}$,
the above implies ${W}^{(0)*}_{MD}~{\preceq}~{W}^{(1)*}_{MD}$,
which is consistent with {\bf [Hyp-2]}.

Thirdly, we consider how much the accuracies differ.
We define the relation ${\cal R}_{obs}$ where 
$obs({W}^{*})~=~Accuracy({W}^{*},{TS})$.
Let ${\cal R}_{obs}^{MD}$ be defined in terms of $Accuracy({W}^{(0)*}_{MD},{TS})$
with a certain ${\epsilon}_{MD}$.
Comparing the graphs in Figure \ref{fig-SGD}(a) and Figure \ref{fig-rel}(a),
we observe, for ${\cal L}_{f}^{PC}$, ${\epsilon}_{PC}$ is about $0.1$.
Contrarily, for ${\cal L}_{f}^{BI}$ from 
Figure \ref{fig-SGD}(b) and Figure \ref{fig-rel}(b),
${\epsilon}_{BI}$ is about $0.4$.
If we choose a threshold to be about $0.2$, the two cases are distinguishable.

Moreover, we define the ${\approx}$ relation for $({\epsilon}{<}{0.2})$.
As we know that ${\cal L}_{f}^{PC}$ is probably correct and
${\cal L}_{f}^{BI}$ is bug-injected, we have followings.
(a) ${W}^{(0)*}_{PC}~{\approx}~{W}^{(1)*}_{PC}$,
and (b) ${W}^{(0)*}_{BI}~{\prec}~{W}^{(1)*}_{BI}$.
These are, indeed, consistent with {\bf [Hyp-3]}.


\subsubsection{Accuracy for Distorted Testing Datasets}

\begin{figure}[!t]
\begin{center}
\includegraphics[width=4.0in]{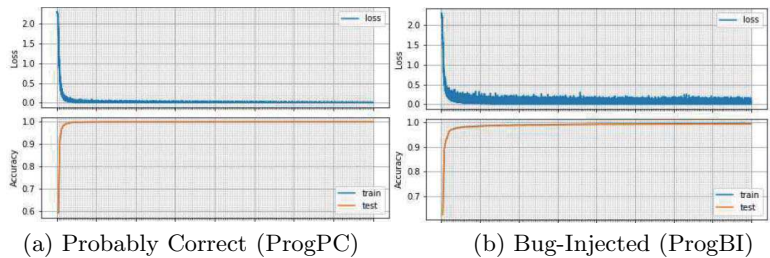}

{(a) Probably Correct (ProgPC)}\hspace*{1.5cm}{(b) Bug-Injected (ProgBI)}

\end{center}
\caption{Loss and Accuracy: Distorted Testing Dataset} \label{fig-g3}
\end{figure}

We generated distorted datasets from the MNIST testing dataset $TS$;
${TS}^{(1)}_{MD} = {T}_{A}(TS, {\cal L}_{f}^{MD}(LS))$.
We, then, checked the accuracy $Accuracy({W}^{(1)}_{MD},{TS}^{(1)}_{MD})$,
whose monitored results are shown in Figure \ref{fig-g3}.
$Accuracy({W}^{(1)}_{MD},{TS}^{(1)}_{MD})$ and
$Accuracy({W}^{(1)}_{MD},{LS}^{(1)}_{MD})$ are not distinguishable,
because both ${LS}^{(1)}_{MD}$ and ${TS}^{(1)}_{MD}$ are constructed
in the same way with ${T}_{A}({\_}, {\cal L}_{f}^{MD}(LS))$
and thus their empirical distributions are the same.
The graphs are consistent again with {\bf [Hyp-1]}.

\section{Discussions}

\subsection{Neuron Coverage}
As explained in Section \ref{subsec-preceq}, the distortion
relation (${\_}{\preceq}{\_}$) between trained learning parameters is
calculated in terms of observer functions.
However, depending on the observer, the resultant
distortion degree may be different.
In an extreme case, a certain observer is not adequate to
differentiate distortions.
A question arises whether such distortion degrees are able to be
measured directly. We will study 
\emph{neuron coverage} \cite{deepxplore2017} whether we can use
it as such a measure.

A neuron is said to be \emph{activated} if its output signal $out$ is larger than 
a given threshold when a set of input signals ${in}_{j}$ is 
presented; {$out~=~{\sigma}({\sum}{w}_{j}{\times}{in}_{j})$}.
The weight values ${w}_{j}$s are constituents of trained ${W}^{*}$.
\emph{Activated Neurons} below refer to a set of neurons that are
activated when a vector data ${\vec{a}}$ is input to a trained learning model
as ${{y}({W}^{*};{\vec{a}})}$.
\[
Neuron~coverage~(NC) ~=~\frac{{\mid}~Activated~Neurons~{\mid}}{{\mid}~Total~Neurons~{\mid}}
\]

\noindent In the above, ${\mid}~{X}~{\mid}$ denotes the size of a set $X$.
Using this neuron coverage as a criterion is motivated by an empirical
observation that 
different input-output pairs result in different degrees of 
neuron coverage \cite{deepxplore2017}.

\subsubsection{Results of Experiment}
\begin{figure}[!t]
\begin{center}
\includegraphics[width=2.2in]{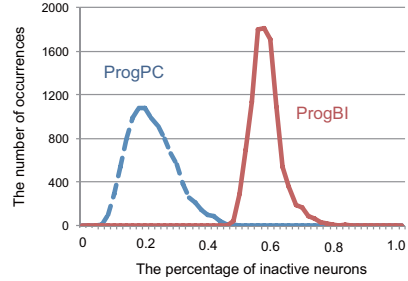}
\end{center}
\caption{Frequencies of Inactive Neurons} \label{fig-ncov}
\end{figure}

We focus on the neurons constituting the hidden layer in our NN learning model.
As its activation function is \emph{ReLU}, we choose $0$ as the threshold.
Figure \ref{fig-ncov} is a graph to show the numbers of input vectors
leading to the chosen percentages of inactive neurons, $(1~-~NC)$.
These input vectors constitute
the MNIST testing dataset $TS$ of the size $10,000$.

According to Figure \ref{fig-ncov}, the graph for the case of ProgPC,
the ratio of inactive neurons is almost $20{\%}$;
namely, $80{\%}$ of neurons in the hidden layer are activated to 
have effects on the classification results. However, the ProgBI graph
shows that about $60{\%}$ of them are inactive and do not contribute
to the results. To put it differently, this difference in the ratios of inactive
neurons implies that the trained learning model ${W}_{BI}^{*}$ of ProgBI is
distorted from ${W}_{PC}^{*}$ of ProgPC, 
${W}_{PC}^{*} {\preceq} {W}_{BI}^{*}$.

From the generation method of the distorted dataset, we have
${LS}~ {\preceq} ~{LS}_{PC}^{(1)}$ and ${LS}~{\preceq}~ {LS}_{BI}^{(1)}$.
${LS}_{PC}^{(1)} {\preceq} {LS}_{BI}^{(1)}$
may also be satisfied, which is in accordance with the visual inspection
of Figure \ref{fig-distorted}.
Furthermore, because of {\bf [Hyp-2]} (Section \ref{subsec-preceq}),
${W}_{PC}^{(1)} ~{\preceq}~ {W}_{BI}^{(1)}$ is true.
It is consistent with the situation shown in Figure \ref{fig-ncov}
in that activated neurons in ${W}_{BI}^{(1)*}$ are fewer
than those in ${W}_{PC}^{(1)*}$.

Figure \ref{fig-rel} can be understood from a viewpoint of the neuron coverage.
The empirical distribution of MNIST testing dataset $TS$ is the same
as that of MNIST training dataset ${LS}$.
Because of the distortion relationships on training datasets,
the distribution of $TS$ is different from those of ${LS}^{(1)}_{MD}$
(${LS}_{PC}^{(1)}$ or ${LS}_{BI}^{(1)}$). 
Moreover,  Figure \ref{fig-rel} shows that
${\mid}Accuracy({W}_{PC}^{(1)*},{TS}) ~-~ Accuracy({W}^{(0)*}_{PC},{TS}){\mid}$ is
smaller than ${\mid}Accuracy({W}_{PC}^{(1)*},{TS}) ~- ~Accuracy({W}_{PC}^{(0)*},{TS}){\mid}$.
Therefore, we see that
the relationship ${LS}_{PC}^{(1)} {\preceq} {LS}_{BI}^{(1)}$ is satisfied.
Because of {\bf [Hyp-2]},  it implies ${W}_{PC}^{(1)*}~ {\preceq}~ {W}_{BI}^{(1)*}$.
Therefore, the difference seen in Figure \ref{fig-rel} is
consistent with the situation shown in Figure \ref{fig-ncov}.

In summary, the neuron coverage would be a good candidate as the metrics
to quantitatively define the distortion degrees of trained learning models.
However, because this view is based on the MNIST dataset experiments only,
further studies are desirable.


\subsection{Test Input Generation} \label{subsec-div}

We will see how the dataset or data generation method in Section \ref{subsec-generation}
is used in software testing.
%
Because the program is categorized as untestable \cite{chenNEW},
Metamorphic Testing (MT) \cite{chen2018} is now a standard practice 
for testing of machine learning programs.
We here indicate that generating an appropriate data is desirable to conduct effective testing.
In the MT framework, given an initial test data $x$, a translation function $T$ generates
a new follow-up test data $T(x)$ automatically.

For testing machine learning software, either 
${\cal L}_{f}$ (whether a training program is faithful implementation of machine learning
algorithms) \cite{nkjm2016}\cite{xie2010} or 
${\cal I}_{f}$ (whether an inference program show acceptable prediction performance against
incoming data), generating a variety of data to show 
\emph{Dataset Diversity} \cite{nkjm2018S} is a key issue.
The function ${T}_{A}$ introduced in Section \ref{subsec-generation} can be used
to generate such a follow-up \emph{dataset} used in the MT framework \cite{nkjm2019I}.
In particular, corner-case testing would be possible by carefully chosen
such a group of biased datasets. 


DeepTest \cite{deeptest2018} employs Data Augmentation methods \cite{krizhevsky2012}
to generate test data. Zhou and Sun \cite{zhou2019} adopts 
generating fuzz to satisfy application-specific properties.
Both works are centered around generating test data for negative testing, but
do not refer to statistical distribution of datasets.

DeepRoad \cite{zhang2018} adopts an approach with
Generative Adversarial Networks (GAN) \cite{goodfellow2014}
to synthesize various weather conditions as driving scenes.
GAN is formulated as a two-player zero-sum game.
Given a dataset whose empirical distribution is ${\rho}^{DS}$,
its Nash equilibrium, solved with Mixed Integer Linear Programing (MILP), 
results in a DNN-based generative model to emit new data to satisfy the relation
$\vec{x}~{\sim}_{i.i.d.}~{\rho}^{DS}$.
Thus, such new data 
preserve characteristics of the original machine learning problem.
Consequently, we regard the GAN-based approach as a method to enlarge
coverage of test scenes within what is anticipated at the training time.


Machine Teaching \cite{zhu2015} is an inverse problem of machine learning, and
is a methodology to obtain a dataset to optimally
derive a \emph{given} trained learning model.
The method is formalized as a two-level optimization problem,
which is generally difficult to solve.
We regard the machine teaching as a method to generate unanticipated
dataset. 
Obtained datasets can be used in negative testing.

Our method uses
an optimization problem with one objective function for generating datasets 
that are not far from what is anticipated, but probably are
biased to build up the dataset diversity.
%


\section{Concluding Remarks} \label{sec-last}
We introduced a notion of distortion degrees which would manifest themselves
as faults and failures in machine learning programs,
and studied the characteristics in terms of neuron coverages.
However, further study would be needed how we rigorously measure
the distortion degrees, which will make it possible to \emph{debug}
programs with the measurement results.
If the measurement is light-weight and can be conducted for
in-operation machine learning systems, we will be able to
diagnose systems at operation time.



\section*{Acknowledgment}
%
The work is supported partially by JSPS KAKENHI Grant Number JP18H03224,
and is partially based on results obtained from a project
commissioned by the NEDO.



\end{document}